\documentclass[journal]{IEEEtran}
\ifCLASSINFOpdf
\else
\fi
\usepackage{hyperref} 
\usepackage{CJKutf8}
\usepackage{graphicx}
\usepackage{booktabs}
\usepackage{enumitem}
\usepackage{bm}
\usepackage{multirow}
\usepackage{subfigure}
\usepackage{amsmath}
\usepackage{amssymb}
\usepackage{color}

\begin{document}
\begin{CJK*}{UTF8}{gbsn}
\title{Pre-Training with Whole Word Masking for Chinese BERT}

% \author{Yiming Cui, Wanxiang Che, Ting Liu, Bing Qin, Ziqing Yang}% <-this % stops a space
\author{{\bf Yiming Cui$^{\dag\ddag}$, Wanxiang Che$^\dag$, Ting Liu$^\dag$, Bing Qin$^\dag$, Ziqing Yang$^{\ddag\S}$}   \\ % \thanks{$^*$Corresponding Author}
{$^\dag$Research Center for Social Computing and Information Retrieval, Harbin Institute of Technology, Harbin, China}\\
{$^\ddag$State Key Laboratory of Cognitive Intelligence, iFLYTEK Research, Beijing, China}\\
{$^\S$iFLYTEK AI Research (Hebei), Langfang, China} \\
{$^\dag$\tt \{ymcui,car,tliu,qinb\}@ir.hit.edu.cn, $^\ddag$$^\S$\tt\{ymcui,zqyang5\}@iflytek.com}\\  
}

% \thanks{Manuscript received April 19, 2005; revised August 26, 2015.}}


\markboth{IEEE/ACM Transactions on Audio, Speech, and Language Processing, November~2021}%
{Cui \MakeLowercase{\textit{et al.}}: Bare Demo of IEEEtran.cls for IEEE Journals}



\maketitle

\begin{abstract}
Bidirectional Encoder Representations from Transformers (BERT) has shown marvelous improvements across various NLP tasks, and its consecutive variants have been proposed to further improve the performance of the pre-trained language models.
In this paper, we aim to first introduce the {\em whole word masking} (wwm) strategy for Chinese BERT, along with a series of Chinese pre-trained language models.
Then we also propose a simple but effective model called MacBERT, which improves upon RoBERTa in several ways.
Especially, we propose a new masking strategy called {\em MLM as correction} (Mac).
To demonstrate the effectiveness of these models, we create a series of Chinese pre-trained language models as our baselines, including BERT, RoBERTa, ELECTRA, RBT, etc.
We carried out extensive experiments on ten Chinese NLP tasks to evaluate the created Chinese pre-trained language models as well as the proposed MacBERT.
Experimental results show that MacBERT could achieve state-of-the-art performances on many NLP tasks, and we also ablate details with several findings that may help future research.
We open-source our pre-trained language models for further facilitating our research community.\footnote{Resources are available: \url{https://github.com/ymcui/Chinese-BERT-wwm}}
\end{abstract}


\newcommand\tcb[1]{\textcolor{blue}{#1}}

% \fi
\IEEEpeerreviewmaketitle

%%%%%%%%%%%%%%%%%%%%%%%%%%%%%%%%%%%%%%%%
\section{Introduction}
\IEEEPARstart{B}{ERT} \cite{devlin-etal-2019-bert} has become enormously popular and has proven to be effective in recent natural language processing studies, which utilizes large-scale unlabeled training data and generates enriched contextual representations.
As we traverse several popular machine reading comprehension benchmarks, such as SQuAD \cite{rajpurkar-etal-2018-know}, CoQA \cite{reddy2019coqa}, QuAC \cite{choi-etal-2018-quac}, NaturalQuestions \cite{kwiatkowski2019natural}, RACE \cite{lai-etal-2017}, we can see that most of the top-performing models are based on BERT and its variants \cite{dai-etal-2019-transformer,zhang2019dual,ran2019option}, demonstrating that the pre-trained language models have become new fundamental components in natural language processing field.

Starting from BERT, the community members have made great and rapid progress on optimizing the pre-trained language models, such as ERNIE \cite{sun2019ernie}, XLNet \cite{yang2019xlnet}, RoBERTa \cite{liu2019roberta}, SpanBERT \cite{joshi2019spanbert}, ALBERT \cite{lan2019albert}, ELECTRA \cite{clark2020electra}, etc. 
However, training Transformer-based \cite{vaswani2017attention} pre-trained language models are not as easy as we used to train word embeddings or other traditional neural networks for learning representations. 
Typically, training a powerful BERT-large model with a 24-layer Transformer and 330 million parameters, to convergence needs high-memory computing devices, such as TPU or TPU Pod, which are very expensive. 
On the other hand, though various pre-trained language models have been released, most of them are based on English, and there are few efforts on building powerful pre-trained language models in other languages. 

To minimize the repetitive work and build baselines for future studies, in this paper, we aim to build Chinese pre-trained language model series and release them to the public for facilitating the research community, as Chinese and English are among the most spoken languages in the world.
We revisit the existing popular pre-trained language models and adjust them to the Chinese language to see whether these models could generalize and perform well in a language other than English.
Besides, we also propose a new pre-trained language model called MacBERT, which replaces the original MLM task into {\bf M}LM {\bf a}s {\bf c}orrection (Mac) task.
MacBERT mainly aims to mitigate the discrepancy of the pre-training and fine-tuning stage in original BERT.
Extensive experiments are conducted on ten popular Chinese NLP datasets, ranging from sentence-level to document-level tasks, such as machine reading comprehension, text classification, etc.
The results show that the proposed MacBERT could give significant gains in most of the tasks against other pre-trained language models, and detailed ablations are given to better examine the composition of the improvements.
The contributions of this paper are listed as follows.

\begin{itemize}
	\item To further accelerate future research on Chinese NLP, we create and release the Chinese pre-trained language model series to our community. Extensive empirical studies are carried out to revisit the performance of these pre-trained language models on various tasks with careful analyses.
	\item We propose a new pre-trained language model called MacBERT that mitigates the gap between the pre-training and fine-tuning stage by masking the word with its similar word, which has proven to be effective on various downstream tasks.
	\item We also create a series of small models, called RBT, to demonstrate how small models perform compared to regular pre-trained language models, which could help utilize them in real-life applications.
\end{itemize}

%%%%%%%%%%%%%%%%%%%%%%%%%%%%%%%%%%%%%%%%
\section{Related Work}
In this section, we revisit the techniques of the representative pre-trained language models in the recent natural language processing field.
The overall comparisons of these models, as well as the proposed MacBERT, are depicted in Table \ref{model-comparison}.
We elaborate on their key components in the following subsections.
\begin{table*}[htbp]
\caption{\label{model-comparison} Comparisons of the pre-trained language models. (AE: Auto-Encoding, AR: Auto-Regressive, T: Token, S: Segment, P: Position, E: Entity, Ph: Phrase, WWM: Whole Word Masking, NM: N-gram Masking, NSP: Next Sentence Prediction, SOP: Sentence Order Prediction, MLM: Masked LM, PLM: Permutation LM, Gen-Dis: Generator-Discriminator, Mac: MLM as correction)}
%\small
\begin{center}
\begin{tabular}{l c c c c c c | c}
\toprule
& \bf BERT & \bf ERNIE & \bf XLNet & \bf RoBERTa & \bf ALBERT & \bf ELECTRA & \bf MacBERT \\
\midrule
Type & AE & AE & AR & AE & AE & AE & AE  \\
Embeddings & T/S/P & T/S/P & T/S/P & T/S/P & T/S/P & T/S/P & T/S/P \\
Masking & T & T/E/Ph & - & T & T & T & WWM/NM \\
LM Task & MLM & MLM & PLM & MLM & MLM & Gen-Dis & Mac \\
Paired Task & NSP & NSP & - & - & SOP & - & SOP \\
\bottomrule
\end{tabular}
\end{center}
\end{table*}

%%%%%%%%%%%%%%
\subsection{BERT}
BERT (Bidirectional Encoder Representations from Transformers) \cite{devlin-etal-2019-bert} has demonstrated its effectiveness in a wide range of natural language processing tasks. 
BERT is designed to pre-train deep bidirectional representations by jointly conditioning on both left and right context in all Transformer layers.
Primarily, BERT consists of two pre-training tasks: Masked Language Model (MLM) and Next Sentence Prediction (NSP).
\begin{itemize}
	\item {\bf MLM}: Randomly masks some of the tokens from the input, and the objective is to predict the original word based only on its context.
	\item {\bf NSP}: To predict whether sentence {\em B} is the next sentence of sentence {\em A}.
\end{itemize}

Later, they further propose a technique called whole word masking (wwm) for optimizing the original masking in the MLM task.
In this setting, instead of randomly selecting WordPiece \cite{wu2016google} tokens to mask, we always mask all of the tokens corresponding to a whole word at once. 
This explicitly forces the model to recover the whole word in the MLM pre-training task instead of just recovering WordPiece tokens \cite{chinese-bert-wwm}, which is much more challenging.
As the whole word masking only affects the masking strategy of the pre-training process, it would not bring additional burdens on downstream tasks.
Moreover, as training pre-trained language models are computationally expensive, they also release all the pre-trained models as well as the source codes, which significantly stimulates the community to have great interests in the research of pre-trained language models.

%%%%%%%%%%%%%%
\subsection{ERNIE}
ERNIE (Enhanced Representation through kNowledge IntEgration) \cite{sun2019ernie} is designed to optimize the masking process of BERT, which includes entity-level masking and phrase-level masking. 
Different from selecting random words in the input, entity-level masking masks the named entities, which are often formed by several words.
Phrase-level masking is to mask consecutive words, which is similar to the N-gram masking strategy \cite{devlin-etal-2019-bert,joshi2019spanbert,Kong2020A}.\footnote{Though N-gram masking was not included in \cite{devlin-etal-2019-bert}, according to their model name in  SQuAD leaderboard, we often admit their credit towards this method.}.

%%%%%%%%%%%%%%
\subsection{XLNet}
\cite{yang2019xlnet} argues that the existing pre-trained language models that are based on auto-encoding, such as BERT, which suffer from the discrepancy of the pre-training and fine-tuning stage because the masking token {\tt [MASK]} never appears in the fine-tuning stage.
To alleviate this problem, XLNet is proposed, which is based on Transformer-XL \cite{dai-etal-2019-transformer}. 
XLNet mainly modifies in two ways. 
The first is to maximize the expected likelihood over all permutations of the factorization order of the input, where they call the Permutation Language Model. 
To achieve this goal, they propose a novel two-stream self-attention mechanism.
Another one is to change the auto-encoding language model into an auto-regressive one, which is similar to the traditional statistical language models.

%%%%%%%%%%%%%%
\subsection{RoBERTa}
RoBERTa (Robustly Optimized BERT Pretraining Approach) \cite{liu2019roberta} aims to adopt original BERT architecture but make much more precise modifications to fully release the power of BERT, which is underestimated in \cite{devlin-etal-2019-bert}. 
They carry out careful comparisons of various components in BERT, including the masking strategies, input format, training steps, etc. 
After thorough evaluations, they come up with several useful conclusions to make BERT more powerful, mainly including 1) training longer with bigger batches and longer sequences over more data; 2) removing the next sentence prediction task and using dynamic masking in MLM task.

%%%%%%%%%%%%%%
\subsection{ALBERT}
ALBERT (A Lite BERT) \cite{lan2019albert} primarily tackles the problems of higher memory consumption and slow training speed of BERT.
ALBERT introduces two techniques for parameter reduction.
The first one is the factorized embedding parameterization, which decomposes the embedding matrix into two small matrices.
The second one is the cross-layer parameter sharing that the Transformer weights are shared across each layer of ALBERT, which significantly reduces the overall parameters.
Besides, they also propose the sentence-order prediction (SOP) task to replace the traditional NSP pre-training task and yield better performances.

\begin{table*}[htbp]
\caption{\label{wwm-example} Examples of different masking strategies. We also include an English example for clarity. Masked tokens are in boldface.}
\scriptsize
\begin{center}
\begin{tabular}{l l l}
        \toprule
        & {\bf Chinese} & {\bf English} \\
        \midrule
	{\bf Original Sentence} & 使用语言模型来预测下一个词的概率。 & we use a language model to predict the probability of the next word.\\
        {\bf + CWS} & 语言~{\bf 模型}~来~{\bf 预测}~下~一个~词~的~{\bf 概率}~。 & - \\
        {\bf + BERT Tokenizer} & 语~言~{\bf 模~型}~来~{\bf 预~测}~下~一~个~词~的~{\bf 概~率}~。 & we~use~a~language~{\bf model}~to~{\bf pre~\#\#di~\#\#ct}~the~{\bf pro~\#\#ba~\#\#bility}~of~the~next~word~.  \\
        \midrule
        {\bf Original Masking} & 语~言~{\bf [M]~型}~来~{\bf [M]~测}~下~一~个~词~的~{\bf 概~率}~。 & we~use~a~language~{\bf [M]}~to~{\bf [M]~\#\#di~\#\#ct}~the~{\bf pro~[M]~\#\#bility}~of~the~next~word~.  \\
        {\bf + WWM} & 语~言~{\bf [M]~[M]}~来~{\bf [M]~[M]}~下~一~个~词~的~{\bf 概~率}~。 & we~use~a~language~{\bf [M]}~to~{\bf [M]~[M]~[M]}~the~{\bf [M]~[M]~[M]}~of~the~next~word~.  \\
        {\bf ++ N-gram Masking} & {\bf [M]~[M]~[M]~[M]}~来~{\bf [M]~[M]}~下~一~个~词~的~{\bf 概~率}~。  & we~use~a~{\bf [M]}~{\bf [M]}~to~{\bf [M]~[M]~[M]}~the~{\bf [M]~[M]~[M]}~{\bf [M]~[M]}~next~word~.  \\
        {\bf +++ Mac Masking} & {\bf 语~法~建~模}~来~{\bf 预~见}~下~一~个~词~的~{\bf 几~率}~。  & we~use~a~{\bf text}~{\bf system}~to~{\bf ca~\#\#lc~\#\#ulate}~the~{\bf po~\#\#si~\#\#bility}~of~the~next~word~.  \\
        \bottomrule
        \end{tabular}

\end{center}
\end{table*}

%%%%%%%%%%%%%%
\subsection{ELECTRA}
ELECTRA (Efficiently Learning an Encoder that Classifiers Token Replacements Accurately) \cite{clark2020electra} employs a new generator-discriminator framework that is similar to generative adversarial net (GAN) \cite{goodfellow-gan-nips2014}.
The generator is typically a small MLM that learns to predict the original words of the masked tokens.
The discriminator is trained to discriminate whether the input token is replaced by the generator, which they call Replaced Token Detection (RTD).
Note that, to achieve efficient training, the discriminator is only required to predict a binary label to indicate ``replacement'', unlike the way of MLM that should predict the exact masked word.
After the pre-training stage, we discard the generator and only use the discriminator for fine-tuning downstream tasks.

%%%%%%%%%%%%%%%%%%%%%%%%%%%%%%%%%%%%%%%%
\section{Chinese Pre-trained Language Models}

While BERT and its variants have achieved significant improvements in various English tasks, we wonder if these models and techniques could generalize well in other languages.
In this section, we illustrate how the existing pre-trained language models are adapted for the Chinese language.
We adopt BERT, RoBERTa, and ELECTRA as well as their variants to create Chinese pre-trained model series, and their effectiveness is shown in Section \ref{sec-results}.
Note that, as these models are all originated from BERT or ELECTRA without changing the nature of the input, no modification should be made to adapt to these models in the fine-tuning stage, which is very flexible for replacing one another.

%%%%%%%%%%%%%%
\subsection{BERT-wwm \& RoBERTa-wwm}
In the original BERT, a WordPiece tokenizer \cite{wu2016google} is used to split the text into WordPiece tokens, where some words are split into several small fragments.
The whole word masking (wwm) mitigates the drawback of masking only a part of the whole word, which is easier for the model to predict.
In Chinese condition, WordPiece tokenizer no longer splits the word into small fragments, as Chinese characters are not formed by alphabet-like symbols.
We use the traditional Chinese Word Segmentation (CWS) tool to split the text into several words.
In this way, we could adopt the whole word masking in Chinese to mask the word instead of individual Chinese characters.
For implementation, we strictly follow the original whole word masking codes and do not change other components, such as the percentage of word masking, etc.
We use LTP \cite{che2010ltp} for Chinese word segmentation to identify the word boundaries.
Note that the whole word masking only affects the selection of the masking tokens in the pre-training stage. 
We still uses WordPiece tokenizer to split the text, which is identical to the original BERT.

Similarly, whole word masking can also be applied on RoBERTa, where the NSP task is not adopted. 
However, we still use a paired input for pre-training, which could be beneficial to the sentence pair classification and reading comprehension tasks.
An example of the whole word masking is depicted in Table \ref{wwm-example}.

%%%%%%%%%%%%%%
\subsection{ELECTRA}
Besides BERT and RoBERTa series, we also explore the ELECTRA model, which adopts a new pre-training framework that consists of a generator and discriminator.
We strictly follow the original implementation as in \cite{clark2020electra}.

%%%%%%%%%%%%%%
\subsection{RBT Series}
Though the aforementioned pre-trained language models are powerful, they are computationally ineffective and hard to adopt in real-life applications.
To make pre-trained models more accessible by community researchers, besides the regular pre-trained language models, we also pre-train several small models, where we call RBT.
Specifically, we use exactly the same training strategy as in training RoBERTa, but we use fewer Transformer layers. 
We train 3-layer, 4-layer, 6-layer RoBERTa-base, denoted as RBT3, RBT4, and RBT6, respectively.
We also train a 3-layer RoBERTa-large, denoted as RBTL3, which has a similar parameter size as RBT6. 
This is designed to compare a wider and shorter model (RBTL3) with a thinner and taller model (RBT6) under a comparable parameter size, which could be useful in the design of future pre-trained language models.

%%%%%%%%%%%%%%
\section{MacBERT}
\begin{figure}[h]
\centering
\includegraphics[width=0.7\columnwidth]{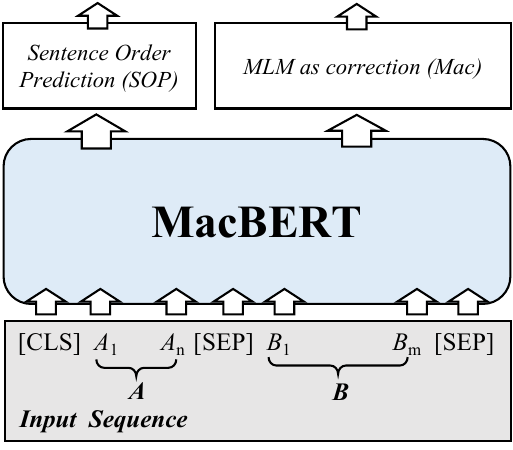} \\
\caption{\label{macbert} Neural architecture of MacBERT.}
\end{figure}

In the previous section, we propose a series of Chinese pre-trained language models. 
In this section, we make the best use of them and propose a novel model called {\bf MacBERT} ({\bf M}LM {\bf a}s {\bf c}orrection {\bf BERT}).
MacBERT shares the similar types of pre-training tasks as BERT with several modifications.
MacBERT consists of two pre-training tasks: MLM as correction, and sentence order prediction.
The overall architecture of MacBERT is depicted in Figure \ref{macbert}.

\subsection{MLM as correction}
Masked Language Model (MLM) is the most important pre-training task in BERT and its variants, which models bidirectional contextual inference ability.
However, as shown in the previous section, the MLM suffers from the `pre-training and fine-tuning' discrepancy, where the artificial tokens in the pre-training stage, such as {\tt [MASK]}, never appear in the real downstream fine-tuning tasks.

To address this issue, we propose a novel pre-training task called MLM as correction (Mac).
In this pre-training task, we do not adopt any pre-defined tokens for masking purposes.
Instead, we transform the original MLM as a text correction task, where the model should correct the wrong word into the correct one, which is much more natural than MLM.
Specifically, in the Mac task, we perform the following modifications on the original MLM.
\begin{itemize}
	\item We use the whole word masking as well as N-gram masking strategies to select candidate tokens for masking, with a percentage of 40\%, 30\%, 20\%, 10\% for word-level unigram to 4-gram. We also notice that a recent work PMI-masking \cite{levine2021pmimasking} is proposed, which optimizes the masking strategy. In this paper, we resort to vanilla N-gram masking and will try PMI-masking in the future.
	\item Instead of masking with {\tt [MASK]} token, which never appears in the fine-tuning stage, we propose to use similar words for the masking purpose. A similar word is obtained by using {\em Synonyms} toolkit \cite{Synonyms:hain2017}, which is based on word2vec \cite{mikolov-etal-2013} similarity calculations. If an N-gram is selected to mask, we find similar words individually. In rare cases, when there is no similar word, we degrade to use random word replacement. Such replacements are restricted to no more than 10\% of all tokens to be masked.
	\item Following previous works, we use a percentage of 15\% input words for masking, where 80\% tokens are replaced with similar words, 10\% tokens are replaced with random words, and keep with original words for the rest of 10\%.
\end{itemize}

\subsection{Sentence Order Prediction}
The original next sentence prediction (NSP) task in BERT is considered to be too easy for the model and proved to be not that effective \cite{liu2019roberta,lan2019albert}.
In this paper, we adopt the sentence order prediction (SOP) task as introduced by ALBERT \cite{lan2019albert}, which is shown to be much more effective than NSP.
The positive samples are created by using two consecutive texts, while the negative ones are created by switching the original order of them.
We ablate these modifications in Section \ref{effect-macbert} to better demonstrate the contributions of each component.

\subsection{Neural Architecture}
Formally, given a pair of sequences $A=\{A_1,\dots,A_n\}$ and $B=\{B_1,\dots,B_m\}$, we first construct the input sequence $X$ by concatenating two sequences.
Then, MacBERT converts $X$ into a contextualized representation $\bm{H}^{(L)}\in\mathbb{R}^{N  \times d}$ through an embedding layer (which consists of word embedding, positional embedding, and token type embedding), and a consecutive $L$-layer transformer, where $N$ is the maximum sequence length, and $d$ is the dimension of hidden layers.
\begin{gather} 
X = {\tt [CLS]} ~{A_1 \dots A_n}~ {\tt [SEP]} ~{B_1 \dots B_m}~ {\tt [SEP]} \\
\bm{H}^{(0)} = \mathbf{Embedding}( X ) \\
\bm{H}^{(i)} = \mathbf{Transformer}(\bm{H}^{(i-1)}),~~i \in \{1,\dots,L\}
\end{gather}

As we only need to predict the positions that are replaced by the Mac task, after getting the contextual representation $\bm{H}^L$, we collect a subset with respect to the replaced positions, forming the replaced representation $\bm{H}^\text{m} \in \mathbb{R}^{k \times d}$, where $k$ is the number of the replaced tokens.
According to the definition of Mac task, $k=\lfloor N \times 15\% \rfloor$.

Then we project $\bm{H}^\text{m}$ into the vocabulary space to predict the probability distributions $\bm{p}$ over the whole vocabulary $\mathbb{V}$.
Following original BERT implementation, we also use word embedding matrix $\bm{W}^\text{e} \in \mathbb{R}^{|\mathbb{V}| \times d}$ to perform the projection, as the embedding and hidden size are identical.
\begin{equation}
	\bm{p}_i = \bm{H}^\text{m}_i \bm{W}^{\text{e}^\top} + \bm{b}
\end{equation}

Then we use the standard cross-entropy loss to optimize the pre-training task.
\begin{equation}\label{equation-ce-loss}
	\mathcal{L} = -\frac{1}{M}\sum_{i=1}^M \bm{y}_i \log \bm{p}_i
\end{equation}

For the SOP task, we directly use the contextual representation of the {\tt [CLS]} token, which is the first component of $\bm{H}$, and project it into the label prediction layer.
\begin{equation}
	\bm{p} = \mathbf{softmax}(\bm{h}_0 \bm{W}^\text{s} + \bm{b}^\text{s})
\end{equation}
where the $\bm{W}^\text{s} \in \mathbb{R}^{d \times 2}$ and $\bm{b}^\text{s} \in \mathbb{R}^{2}$ are the weight matrix and bias.
Then we also use the cross-entropy loss to optimize the pre-training task (similar to Equation \ref{equation-ce-loss}).
Finally, the overall training loss is the combination of the Mac and SOP task.
\begin{equation}
	\mathcal{L} = \mathcal{L}_{mac} + \mathcal{L}_{sop}
\end{equation}

\begin{table*}[htbp]
\caption{\label{comparison-base} Training details of Chinese pre-trained language models.}
\begin{center}
\begin{tabular}{l c c c c c c}
\toprule
 	& \bf BERT & \bf BERT-wwm & \bf RoBERTa-wwm & \bf RBT & \bf ELECTRA & \bf MacBERT \\
\midrule
Word \#			& 0.4B & 5.4B  & 5.4B & 5.4B & 5.4B & 5.4B \\
Vocab \#			& 21,128 & 21,128 & 21,128 & 21,128 & 21,128 & 21,128 \\
Hidden Activation	& GeLU & GeLU  & GeLU & GeLU & GeLU & GeLU \\
Optimizer 			& AdamW & LAMB & AdamW & AdamW & AdamW & LAMB \\
Training Steps (base/large)		& ? & 2M & 1M / 2M & 1M & 1M / 2M & 1M / 2M \\
Initial Checkpoint (base) & random & BERT & BERT & RoBERTa & random & BERT \\
\bottomrule
\end{tabular}
\end{center}
\end{table*}

%%%%%%%%%%%%%%%%%%%%%%%%%%%%%%%%%%%%%%%%
\section{Experimental Setups}
%%%%%%%%%%%%%%
\subsection{Data Processing}
We use Wikipedia dump\footnote{https://dumps.wikimedia.org/zhwiki/latest/} (as of March 25, 2019), and pre-process with {\tt WikiExtractor.py} as suggested by \cite{devlin-etal-2019-bert}, resulting in 1,307 extracted files. 
We use both Simplified and Traditional Chinese in this dump and do not convert the Traditional Chinese portion into Simplified one. We demonstrate the effectiveness in the Traditional Chinese task in Section \ref{sec-results-mrc}.
After cleaning the raw text, such as removing {\tt html} tags and separating the document, we obtain about 0.4B words.
As Chinese Wikipedia data is relatively small, besides Chinese Wikipedia, we also use extended training data for training these pre-trained language models (mark with {\tt ext} in the model name).
The in-house collected extended data contains encyclopedia, news, and question answering web, which has 5.4B words and is over ten times bigger than the Chinese Wikipedia.
Note that we always use extended data for MacBERT and omit the {\tt ext} mark.
In order to identify the boundary of Chinese words for whole word masking, we use LTP \cite{che2010ltp} for Chinese word segmentation.
We use official {\tt create\_pretraining\_data.py} provided by \cite{devlin-etal-2019-bert} to convert the raw input text to the pre-training examples.

\subsection{Setups for Pre-Trained Language Models}
To better acquire the knowledge from the existing pre-trained language model, we did NOT train our base-level model from scratch but the official Chinese BERT-base, inheriting its vocabulary and weight.
However, for the large-level model, we have to train from scratch but still using the same vocabulary provided by the base-level model.
The base-level model is a 12-layer transformer with a hidden dimension of 768, while the large-level model is a 24-layer transformer with a hidden dimension of 1024.

For training BERT series, we adopt the scheme of training on a maximum sequence length of 128 tokens then on 512, suggested by \cite{devlin-etal-2019-bert}.
However, we empirically found that this results in insufficient adaptation for the long-sequence tasks, such as reading comprehension.
In this context, for models other than BERT, we directly use a maximum length of 512 throughout the pre-training process, which is adopted in \cite{liu2019roberta}.
For smaller batch sizes, we adopt the original \textsc{Adam} \cite{kingma2014adam} with weight decay optimizer in BERT for optimization, and use \textsc{LAMB} optimizer \cite{you2019reducing} for better scalability in larger batch size.
The pre-training was either done on a single Google Cloud TPU\footnote{https://cloud.google.com/tpu/} v3-8 (equals to a single TPU) or TPU Pod v3-32 (equals to 4 TPUs), depending on the magnitude of the model. 
Specifically, for MacBERT-large, we trained for 2M steps with a batch size of 512 and an initial learning rate of 1e-4.

The training details are shown in Table \ref{comparison-base}. 
For clarity, we do not list `{\tt ext}' models, where the other parameters are the same as the one that is not trained on extended data.

%%%%%%%%%%%%%%
\subsection{Setups for Fine-tuning Tasks}
To thoroughly test these pre-trained language models, we carry out extensive experiments on various natural language processing tasks, covering a wide spectrum of text length, i.e., from sentence-level to document-level. Task details are shown in Table \ref{hyper}.
Specifically, we choose the following ten popular Chinese datasets.
\begin{itemize}[leftmargin=*]
	\item {\bf Machine Reading Comprehension (MRC)}: CMRC 2018 \cite{cui-emnlp2019-cmrc2018}, DRCD \cite{shao2018drcd}, CJRC \cite{duan2019cjrc}.
	\item {\bf Single Sentence Classification (SSC)}: ChnSentiCorp \cite{tan2008empirical}, THUCNews \cite{li2007scalable}, TNEWS\cite{clue}.
	\item {\bf Sentence Pair Classification (SPC)}: XNLI \cite{conneau2018xnli}, LCQMC \cite{liu2018lcqmc}, BQ Corpus \cite{chen-etal-2018-bq}, OCNLI\cite{ocnli}.
\end{itemize}

\begin{table}[tbp]
\caption{\label{hyper} Data statistics and hyper-parameter settings for different fine-tuning tasks.}
\begin{center}
\begin{tabular}{l c c c | c c c}
\toprule
\bf Dataset  & \bf MaxLen & \bf Epoch & \bf LR & \bf Train & \bf Dev & \bf Test \\
\midrule
CMRC 2018   		& 512 & 2 & 3e-5 & 10K & 3.2K & 4.9K  \\
DRCD 			& 512 & 2 & 3e-5 & 27K & 3.5K & 3.5K  \\
CJRC 			& 512 & 2 & 4e-5 & 10K & 3.2K & 3.2K  \\
\midrule
ChnSentiCorp 	& 256 & 3 & 2e-5 & 9.6K & 1.2K & 1.2K \\
THUCNews 		& 512 & 3 & 2e-5 & 50K & 5K & 10K  \\
TNEWS		& 128 & 3 & 2e-5 & 53.3K & 10K & 10K \\
\midrule
XNLI			& 128 & 2 & 3e-5 & 392K & 2.5K & 5K  \\
LCQMC 			& 128 & 3 & 2e-5 & 240K & 8.8K & 12.5K  \\
BQ Corpus	  	& 128 & 3 & 3e-5 & 100K & 10K & 10K  \\
OCNLI 		& 128 & 3 & 2e-5 & 56K & 3K & 3K \\
\bottomrule
\end{tabular}
\end{center}
\end{table}

\begin{table*}[htbp]
\caption{\label{result-cmrc2018} Results on CMRC 2018 (Simplified Chinese) and DRCD. The average scores of 10 independent runs are depicted in brackets. Overall best performances are depicted in boldface (base-level and large-level are marked individually).}
\begin{center}
\begin{tabular}{l c c c c c c | c c c c}
\toprule
 & \multicolumn{6}{c}{\centering \bf CMRC 2018} & \multicolumn{4}{c}{\centering \bf DRCD} \\
 & \multicolumn{2}{c}{\centering \bf Dev} & \multicolumn{2}{c}{\centering \bf Test} & \multicolumn{2}{c}{\centering \bf Challenge} & \multicolumn{2}{c}{\centering \bf Dev} & \multicolumn{2}{c}{\centering \bf Test} \\
& \bf EM & \bf F1 & \bf EM & \bf F1  & \bf EM & \bf F1  & \bf EM & \bf F1  & \bf EM & \bf F1  \\
\midrule
BERT   			& 65.5 \tiny(64.4) & 84.5 \tiny(84.0) & 70.0 \tiny(68.7) & 87.0 \tiny(86.3)  & 18.6 \tiny(17.0) & 43.3 \tiny(41.3) & 83.1 \tiny(82.7) & 89.9 \tiny(89.6) & 82.2 \tiny(81.6) & 89.2 \tiny(88.8) \\
BERT-wwm     	& 66.3 \tiny(65.0) & 85.6 \tiny(84.7) & 70.5 \tiny(69.1) & 87.4 \tiny(86.7)  & 21.0 \tiny(19.3) & 47.0 \tiny(43.9) & 84.3 \tiny(83.4) & 90.5 \tiny(90.2) & 82.8 \tiny(81.8) & 89.7 \tiny(89.0) \\
BERT-wwm-ext	& 67.1 \tiny(65.6) & 85.7 \tiny(85.0) & 71.4 \tiny(70.0) & 87.7 \tiny(87.0) & 24.0 \tiny(20.0) & 47.3 \tiny(44.6) & 85.0 \tiny(84.5) & 91.2 \tiny(90.9) & 83.6 \tiny(83.0) & 90.4 \tiny(89.9)  \\
RoBERTa-wwm-ext & 67.4 \tiny(66.5) & 87.2 \tiny(86.5) & 72.6 \tiny(71.4) & 89.4 \tiny(88.8) & 26.2 \tiny(24.6) & 51.0 \tiny(49.1) & 86.6 \tiny(85.9) & 92.5 \tiny(92.2) & 85.6 \tiny(85.2) & 92.0 \tiny(91.7)  \\
ELECTRA-base & 68.4 \bf\tiny(68.0) & 84.8 \tiny(84.6) & 73.1 \bf\tiny(72.7) & 87.1 \tiny(86.9) & 22.6 \tiny(21.7) & 45.0 \tiny(43.8) & 87.5 \tiny(87.0) & 92.5 \tiny(92.3) & 86.9 \tiny(86.6) & 91.8 \tiny(91.7) \\
\bf MacBERT-base & {\bf 68.5} \tiny(67.3) & \bf 87.9 \tiny(87.1) & {\bf 73.2} \tiny(72.4) & \bf 89.5 \tiny(89.2) & \bf 30.2 \tiny(26.4) & \bf 54.0 \tiny(52.2) & \bf 89.4 \tiny(89.2) & \bf 94.3 \tiny(94.1) & \bf 89.5 \tiny(88.7) & \bf 93.8 \tiny(93.5) \\
\midrule
ELECTRA-large & 69.1 \tiny(68.2) & 85.2 \tiny(84.5) & 73.9 \tiny(72.8) & 87.1 \tiny(86.6) & 23.0 \tiny(21.6) & 44.2 \tiny(43.2) & 88.8 \tiny(88.7) & 93.3 \tiny(93.2) & 88.8 \tiny(88.2) & 93.6 \tiny(93.2) \\
RoBERTa-wwm-ext-large & 68.5 \tiny(67.6) & 88.4 \tiny(87.9) & 74.2 \tiny(72.4) & 90.6 \tiny(90.0) & 31.5 \bf \tiny(30.1) & 60.1 \tiny(57.5)  & 89.6 \tiny(89.1) & 94.8 \tiny(94.4) & 89.6 \tiny(88.9) & 94.5 \tiny(94.1) \\
\bf MacBERT-large 		& \bf 70.7 \tiny(68.6) & \bf 88.9 \tiny(88.2) & \bf 74.8 \tiny(73.2) & \bf 90.7 \tiny(90.1) & {\bf 31.9} \tiny(29.6) & \bf 60.2 \tiny(57.6) & \bf 91.2 \tiny(90.8) & \bf 95.6 \tiny(95.3) & \bf 91.7 \tiny(90.9) & \bf 95.6 \tiny(95.3) \\
\bottomrule
\end{tabular}
\end{center}
\end{table*}

In order to make a fair comparison, for each dataset, we keep the same hyper-parameters (such as maximum length, warm-up steps, etc.) and only tune the initial learning rate from 1e-5 to 5e-5 for each task.
Note that the initial learning rates are tuned on the original Chinese BERT, and it would be possible to achieve another gain by tuning the learning rate individually. 
We run the same experiment ten times to ensure the reliability of the results.
The best initial learning rate is determined by selecting the best average development set performance.
We report the maximum and average scores to both evaluate the peak and average performance.
Except for TNEWS and OCNLI, where the test sets are not publicly available, we report both development and test set results.

For all models except for ELECTRA, we use the same initial learning rate setting for each task, as depicted in Table \ref{hyper}.
For ELECTRA models, we use a universal initial learning rate of 1e-4 for base-level models and 5e-5 for large-level models as suggested in \cite{clark2020electra}.

As the pre-training data are quite different among various existing Chinese pre-trained language models, such as ERNIE \cite{sun2019ernie}, ERNIE 2.0 \cite{sun2019ernie2}, NEZHA \cite{wei2019nezha}, we only compare BERT \cite{devlin-etal-2019-bert}, BERT-wwm, BERT-wwm-ext, RoBERTa-wwm-ext, RoBERTa-wwm-ext-large, ELECTRA, along with our MacBERT to ensure relatively fair comparisons among different models, where all models are trained by ourselves except for the original Chinese BERT \cite{devlin-etal-2019-bert}.
We carried out experiments under TensorFlow framework \cite{abadi2016tensorflow} with slight modifications to the fine-tuning scripts\footnote{https://github.com/google-research/bert} provided by \cite{devlin-etal-2019-bert} to better adapt to Chinese tasks.

%%%%%%%%%%%%%%%%%%%%%%%%%%%%%
\section{Results}\label{sec-results}
%%%%%%%%%%%%%%
\subsection{Machine Reading Comprehension}\label{sec-results-mrc}
Machine Reading Comprehension (MRC) is a representative document-level modeling task that requires to answer the questions based on the given passages.
We mainly test these models on three datasets: CMRC 2018, DRCD, and CJRC.
\begin{itemize}
	\item {\bf CMRC 2018}: A span-extraction machine reading comprehension dataset, which is similar to SQuAD \cite{rajpurkar-etal-2016} that extract a passage span for the given question.
	\item {\bf DRCD}: This is also a span-extraction MRC dataset but in Traditional Chinese. 
	\item {\bf CJRC}: Similar to CoQA \cite{reddy2019coqa}, which has yes/no questions, no-answer questions, and span-extraction questions. The data is collected from Chinese law judgment documents. Note that we only use {\tt small-train-data.json} for training. 
 \end{itemize}

\begin{table}[t]
\caption{\label{result-cjrc} Results on CJRC. }
\begin{center}
\begin{tabular}{l c c c c }
\toprule
\multirow{2}*{\bf CJRC} & \multicolumn{2}{c}{\centering \bf Dev} & \multicolumn{2}{c}{\centering \bf Test} \\
& \bf EM & \bf F1 & \bf EM & \bf F1 \\
\midrule
BERT   			& 54.6 \tiny(54.0) & 75.4 \tiny(74.5) & 55.1 \tiny(54.1) & 75.2 \tiny(74.3) \\
BERT-wwm    	& 54.7 \tiny(54.0) & 75.2 \tiny(74.8) & 55.1 \tiny(54.1) & 75.4 \tiny(74.4) \\
BERT-wwm-ext	& 55.6 \tiny(54.8) & 76.0 \tiny(75.3) & 55.6 \tiny(54.9) & 75.8 \tiny(75.0) \\
RoBERTa-wwm-ext & 58.7 \tiny(57.6) & 79.1 \tiny(78.3) & 59.0 \tiny(57.8) & 79.0 \tiny(78.0) \\
ELECTRA-base & 59.0 \tiny(58.1) & 79.4 \tiny(78.5) & 59.3 \tiny(58.2) & 79.4 \tiny(78.3) \\
\bf MacBERT-base & \bf 60.4 \tiny(59.5) & \bf 80.3 \tiny(79.2) & \bf 60.3 \tiny(59.3) & \bf 79.8 \tiny(79.0)  \\
\midrule
ELECTRA-large & 61.9 \tiny(60.8) & 82.1 \tiny(81.2) & 62.3 \tiny(61.2) & 82.0 \tiny(80.7) \\ 
RoBERTa-wwm-ext-L & 62.1 \tiny(61.1) & \bf 82.4 \tiny(81.6) & 62.4 \tiny(61.4) & 82.2 \tiny(81.0) \\
\bf MacBERT-large & \bf 62.4 \tiny(61.3) & 82.3 \tiny(81.4) & \bf 62.9 \tiny(61.6) & \bf 82.5 \tiny(81.1) \\
\bottomrule
\end{tabular}
\end{center}
\end{table}

The results are depicted in Table \ref{result-cmrc2018} and \ref{result-cjrc}.
Using additional pre-training data results in further improvement, as shown in the comparison between BERT-wwm and BERT-wwm-ext.
This is why we use extended data for RoBERTa, ELECTRA, and MacBERT.
Moreover, the proposed MacBERT yields significant improvements on all reading comprehension datasets.
It is worth mentioning that our MacBERT-large could achieve a state-of-the-art F1 of 60\% on the challenge set of CMRC 2018, which requires deeper text understanding.

Also, it should be noted that though DRCD is a traditional Chinese dataset, training with additional large-scale simplified Chinese could also have a great positive effect.
As simplified and traditional Chinese share many identical characters, using a powerful pre-trained language model with only a few traditional Chinese data could also bring improvements without converting traditional Chinese characters into simplified ones.

Regarding CJRC, where the text is written in professional ways regarding Chinese laws, BERT-wwm shows moderate improvement over BERT but not that salient, indicating that further domain adaptation is needed for the fine-tuning tasks on non-general domains.
However, increasing general pre-training data results in improvement, suggesting that when there is not enough domain data, we could also use large-scale general data as a remedy.

\begin{table}[htbp]
\caption{\label{result-spm} Results on single sentence classification tasks: ChnSentiCorp, THUCNews and TNEWS. `R' stands for RoBERTa, `E' stands for ELECTRA, `M' stands for `MacBERT'.}
\begin{center}
\begin{tabular}{l cc | cc | c }
\toprule
\bf & \multicolumn{2}{c}{\centering \bf ChnSentiCorp} & \multicolumn{2}{c}{\centering \bf THUCNews} & \bf TNEWS \\
 & \bf Dev & \bf Test & \bf Dev & \bf Test  & \bf Dev  \\
\midrule
BERT     		& 94.7 \tiny(94.3) & 95.0 \tiny(94.7) & 97.7 \tiny(97.4) & \bf 97.8 \tiny(97.6) & 56.3 \tiny(56.1) \\
BERT-w     	& 95.1 \tiny(94.5) & 95.4 \bf\tiny(95.0) & 98.0 \tiny(97.6) & \bf 97.8 \tiny(97.6) & 56.5 \tiny(56.3)  \\
BERT-w-e	 	& {\bf 95.4} \tiny(94.6) & 95.3 \tiny(94.8) & 97.7 \tiny(97.5) & 97.7 \tiny(97.5) & 57.0 \tiny(56.6) \\
R-base & 94.9 \tiny(94.6) & {\bf 95.6} \tiny(94.9) & {\bf 98.3} \tiny(97.9) & 97.8 \tiny(97.5) & {\bf 57.4} \tiny(56.9) \\
E-base	& 93.8 \tiny(93.0) & 94.5 \tiny(93.5) & 98.1 \tiny(97.9) & 97.8 \tiny(97.5) & 56.1 \tiny(55.7) \\  
\bf M-base & 95.2 \bf\tiny(94.8) & {\bf 95.6} \tiny(94.9) & 98.2 \bf \tiny(98.0) &  97.7 \tiny(97.5)  & \bf 57.4 \tiny(57.1) \\
\midrule
E-large 	& 95.2 \tiny(94.6) & 95.3 \tiny(94.8) & 98.2 \bf\tiny(97.8) & 97.8 \tiny(97.6) & 57.2 \tiny(56.9) \\
R-large & {\bf 95.8} \tiny(94.9) & 95.8 \tiny(94.9) & {\bf 98.3} \tiny(97.7) & 97.8 \tiny(97.6) & 58.8 \tiny(58.4)  \\
\bf M-large & 95.7 \bf \tiny(95.0) & {\bf 95.9} \bf\tiny(95.1) & 98.1 \bf\tiny(97.8) & \bf 97.9 \tiny(97.7)  & \bf 59.0 \tiny(58.8) \\
\bottomrule
\end{tabular}
\end{center}
\end{table}

\begin{table*}[htbp]
\caption{\label{result-spm} Results on sentence pair classification tasks: XNLI, LCQMC, BQ Corpus, and OCNLI. }
\begin{center}
\begin{tabular}{l cc | cc | cc | c}
\toprule
\bf & \multicolumn{2}{c}{\centering \bf XNLI} & \multicolumn{2}{c}{\centering \bf LCQMC} & \multicolumn{2}{c}{\centering \bf BQ Corpus} & \bf OCNLI \\
 & \bf Dev & \bf Test  & \bf Dev & \bf Test & \bf Dev & \bf Test & \bf Dev  \\
\midrule
BERT     		& 77.8 \tiny(77.4) & 77.8 \tiny(77.5) & 89.4 \tiny(88.4) & 86.9 \tiny(86.4) 	& 86.0 \tiny(85.5) 	& 84.8 \tiny(84.6) & 74.6 \tiny(74.2)  \\
BERT-wwm     	& 79.0 \tiny(78.4) & 78.2 \tiny(78.0) & 89.4 \tiny(89.2) & 87.0 \tiny(86.8)  & 86.1 \bf\tiny(85.6) 	& 85.2 \bf\tiny(84.9) & 74.6 \tiny(74.3) \\
BERT-wwm-ext	 & 79.4 \tiny(78.6) & 78.7 \tiny(78.3)  & 89.6 \tiny(89.2) & 87.1 \tiny(86.6) & {\bf 86.4} \tiny(85.5)  & {\bf 85.3} \tiny(84.8) & 76.0 \tiny(75.3)  \\
RoBERTa-wwm-ext & 80.0 \tiny(79.2) & 78.8 \tiny(78.3)  & 89.0 \tiny(88.7) & 86.4 \tiny(86.1) & 86.0 \tiny(85.4) & 85.0 \tiny(84.6) & 76.5 \tiny(76.0)   \\
ELECTRA-base 	& 77.9 \tiny(77.0) & 78.4 \tiny(77.8) & \bf 90.2 \tiny(89.8) & \bf 87.6 \tiny(87.3) & 84.8 \tiny(84.7) & 84.5 \tiny(84.0) & 76.1 \tiny(75.8) \\  
\bf MacBERT-base & \bf 80.3 \tiny(79.7) & \bf 79.3 \tiny(78.8) & 89.5 \tiny(89.3) & 87.0 \tiny(86.5) & 86.0 \tiny(85.5) & 85.2 \bf\tiny(84.9) & \bf 77.0 \tiny(76.5) \\
\midrule
ELECTRA-large 	& 81.5 \tiny(80.8) & 81.0 \bf\tiny(80.9) & \bf 90.7 \tiny(90.4) & 87.3 \bf\tiny(87.2) & \bf 86.7 \tiny(86.2) & 85.1 \tiny(84.8) & 78.8 \tiny(78.4)  \\
RoBERTa-wwm-ext-large & 82.1 \tiny(81.3) & 81.2 \tiny(80.6)  & 90.4 \tiny(90.0) & 87.0 \tiny(86.8) & 86.3 \tiny(85.7) & {\bf 85.8} \tiny(84.9) & 78.5 \tiny(78.2)  \\
\bf MacBERT-large & \bf 82.4 \tiny(81.8) & {\bf 81.3} \tiny(80.6) & 90.6 \tiny(90.3) & {\bf 87.6} \tiny(87.1) & 86.2 \bf \tiny(85.7) & 85.6 \bf \tiny(85.0) & \bf 79.0 \tiny(78.7)  \\
\bottomrule
\end{tabular}
\end{center}
\end{table*}

%%%%%%%%%%%%%%
\subsection{Single Sentence Classification}
For the single sentence classification tasks, we select ChnSentiCorp, THUCNews, and TNEWS datasets.
We use the ChnSentiCorp for evaluating sentiment classification, where the text should be classified into either a positive or negative label.
THUCNews is a dataset that contains news in different genres, where the text is typically very long.
In this paper, we use a version that contains 50K news in 10 domains (evenly distributed), including sports, finance, technology, etc.\footnote{https://github.com/gaussic/text-classification-cnn-rnn}
TNEWS is a short text classification task consisting of news titles and keywords. TNEWS requires to classify into one of 15 classes. 
The results show that MacBERT could give moderate improvements over baselines in ChnSentiCorp and THUCNews, as these datasets have already reached high accuracies.
In TNEWS, we can see that our MacBERT yields consistent improvements across base-level and large-level PLMs.

%%%%%%%%%%%%%%
\subsection{Sentence Pair Classification}
For sentence pair classification tasks, we use XNLI data (Chinese portion), Large-scale Chinese Question Matching Corpus (LCQMC), BQ Corpus, and OCNLI, which require to input two sequences and predict their relations.

In XNLI and OCNLI, we can see that MacBERT yields relatively consistent and significant improvements over baselines.
However, MacBERT only shows moderate improvements on LCQMC and BQ Corpus, with a slight improvement on the average score, but the peak performance is not as good as RoBERTa-wwm-ext-large. 
We suspect that these tasks are less sensitive to the subtle difference of the input than the reading comprehension tasks. 
As sentence pair classification only needs to generate a unified representation of the whole input and thus results in a moderate improvement.

We also noticed that the improvements are bigger in MRC tasks than classification tasks, while it might attribute to the masking strategy.
In MRC tasks, the models should identify the exact answer span in the passage. 
In MacBERT, each word of N-gram is either replaced by its synonym or a random word, and thus each word can be easily identified, which forces the model to learn the word boundaries. 

Another observation is that MacBERT-base generally yields larger improvements than MacBERT-large. This might be caused by two reasons. Firstly, MacBERT-base is initialized by BERT-base, which could benefit from the knowledge in BERT-base and avoid the cold-starting issue. Secondly, the results of large-level PLMs are generally higher than those of base-level PLMs, and thus getting a higher score is much difficult than base-level PLMs.

%%%%%%%%%%%%%%
\subsection{Results on Small Models}
We also build a series of small models, namely RBT, built on either RoBERTa-base or RoBERTa-large models.
The experimental results are shown in Table \ref{rbt-results}.
Small models perform worse than the general models (base-level, large-level), because they use fewer parameters.
As we can see that the performance drops in classification tasks are smaller than the reading comprehension tasks, indicating that it is possible to sacrifice minor performance to obtain a faster and smaller model, which could be beneficial for real-life applications.
Also, by comparing RBTL3 and RBT6, which have similar parameter sizes, we can see that RBT6 substantially outperforms RBTL3, which indicates that a thin-and-tall model usually outperforms a wide-and-short model.
These observations could be helpful in future model design for real-life applications.

\begin{table*}[htbp]
\caption{\label{rbt-results} Results on RBT series, which are built on RoBERTa-large (RoBERTa-wwm-ext-large) and RoBERTa-base (RoBERTa-wwm-ext).}
\begin{center}
\begin{tabular}{l c | cc | cc | cc | c c | c c c | c}
\toprule
\multirow{2}*{\bf System} & \multirow{2}*{\bf Params} & \multicolumn{2}{c}{\centering \bf CMRC 2018} & \multicolumn{2}{c}{\centering \bf DRCD}  & \multicolumn{2}{c}{\centering \bf CJRC} & {\bf CSC} & {\bf THUC} & {\bf XNLI} & {\bf LC} &  {\bf BQ} & \multirow{2}*{\bf AVG}  \\
& & \bf EM & \bf F1 & \bf EM & \bf F1 & \bf EM & \bf F1  & \bf ACC & \bf ACC & \bf ACC & \bf ACC & \bf ACC \\
\midrule
RoBERTa-large & 324M & 74.2 & 90.6 & 89.6 & 94.5 & 62.4 & 82.2 & 95.8 & 97.8 & 81.2 & 87.0 & 85.8 & 86.79 \\
RoBERTa-base & 102M & 72.6 & 89.4 & 85.6 & 92.0 & 59.0 & 79.0 & 95.6 & 97.8 & 78.8 & 86.4 & 85.0 & 85.30 \\
RBTL3 	& 61M & 63.3 & 83.4 & 77.2 & 85.6 & 64.6 & 74.9 & 94.2 & 97.8 & 74.0 & 85.1 & 83.6 & 82.40 \\
RBT3 	& 38M & 62.2 & 81.8 & 75.0 & 83.9 & 63.5 & 73.7 & 92.8 & 97.5 & 72.3 & 85.1 & 83.3 & 81.38 \\
RBT4 	& 45M & 65.0 & 83.9 & 78.7 & 86.7 & 65.5 & 75.3 & 93.8 & 97.7 & 74.2 & 85.7 & 83.7 & 82.83 \\
RBT6 	& 60M & 68.3 & 84.4 & 83.9 & 90.2 & 69.1 & 78.8 & 95.3 & 97.8 & 76.2 & 86.6 & 84.2 & 84.68 \\
\bottomrule
\end{tabular}
\end{center}
\end{table*}

%%%%%%%%%%%%%%%%%%%%%%%%%%%%%%%%%%%%%%%%
\section{Discussion}
Based on the experimental results, we can see that these pre-trained language models also yield significant improvements over traditional BERT in Chinese tasks, indicating their effectiveness and generalizability.
While our models achieve significant improvements on various Chinese tasks, we wonder where the essential components of the improvements from.
To this end, we carry out detailed ablations on MacBERT to demonstrate its effectiveness,
and we also compare the claims of the existing pre-trained language models in English to see if their modification still holds true in another language.

%%%%%%%%%%%%%%
\subsection{Effectiveness of MacBERT}\label{effect-macbert}
We carry out detailed ablations to examine the contributions of each component in MacBERT. 
The results are shown in Table \ref{ablations}. 
\begin{table*}[htbp]
\caption{\label{ablations} Ablations of MacBERT-large on different fine-tuning tasks.}
\begin{center}
\begin{tabular}{l cc | cc | cc | c c | c c c | c}
\toprule
\multirow{2}*{\bf System} & \multicolumn{2}{c}{\centering \bf CMRC 2018} & \multicolumn{2}{c}{\centering \bf DRCD}  & \multicolumn{2}{c}{\centering \bf CJRC} & {\bf CSC} & {\bf THUC} & {\bf XNLI} & {\bf LC} &  {\bf BQ} & \multirow{2}*{\bf AVG}  \\
& \bf EM & \bf F1 & \bf EM & \bf F1 & \bf EM & \bf F1  & \bf ACC & \bf ACC & \bf ACC & \bf ACC & \bf ACC \\
\midrule
MacBERT-large		& 74.8 & 90.7 & 91.7 & 95.6 & 62.9 & 82.5 & 95.9 & 97.9 & 81.3 & 87.6 & 85.6 & 87.18 \\
SOP $\rightarrow$ NSP	& 74.5 & 90.6 & 91.5 & 95.5 & 62.4 & 82.3 & 96.0 & 97.8 & 81.2 & 87.4 & 85.2 & 87.00 \\
w/o SOP 				& 74.4 & 90.6 & 91.0 & 95.4 & 62.2 & 82.1 & 95.8 & 97.8 & 81.1 & 87.4 & 85.2 & 86.89 \\
\midrule
w/o Mac	& 74.2 & 90.1 & 91.2 & 95.4 & 62.2 & 82.3 & 95.7 & 97.8 & 81.2 & 87.4 & 85.3 & 86.88  \\
w/o NM 	& 74.0 & 89.8 & 90.9 & 95.1 & 62.1 & 82.0 & 95.9 & 97.9 & 81.3 & 87.5 & 85.6 & 86.89 \\
RoBERTa-large 	& 74.2 & 90.6 & 89.6 & 94.5 & 62.4 & 82.2 & 95.8 & 97.8 & 81.2 & 87.0 & 85.8 & 86.79 \\
\bottomrule
\end{tabular}
\end{center}
\end{table*}

The overall average scores are obtained by averaging the test scores of each task (EM and F1 metrics are averaged before the overall averaging).
From a general view, removing any component in MacBERT results in a decline in the average performance, suggesting that all modifications contribute to the overall improvements.
Specifically, the most effective modifications are the N-gram masking and similar word replacement, which are the modifications on the masked language model task.
When we compare N-gram masking and similar word replacement, we could see clear pros and cons, where N-gram masking seems to be more effective in text classification tasks, and the performance of reading comprehension tasks seems to benefit more from the similar word replacement task.
Combining these two tasks could compensate for each other and have a better performance on both genres.

The NSP task does not show as much importance as the MLM task, demonstrating that it is much more important to design a better MLM task to fully unleash the text modeling power.
Also, we compared the next sentence prediction \cite{devlin-etal-2019-bert} and sentence order prediction \cite{lan2019albert} task to better judge which one is much powerful.
The results show that the sentence order prediction task indeed shows better performance than the original NSP, though it is not that salient.
The SOP task requires identifying the correct order of the two sentences rather than using a random sentence, which is much easy for the machine to identify. 
Removing the SOP task results in noticeable declines in reading comprehension tasks compared to the text classification tasks, which suggests that it is necessary to design an NSP-like task to learn the relations between two segments (for example, passage and question in reading comprehension task).

%%%%%%%%%%%%%%
\subsection{Investigation on MLM Task}
As illustrated in the previous section, the dominant pre-training task is the masked language model and its variants.
The masked language model task relies on two sides: 1) the selection of the tokens to be masked, and 2) the replacement of the selected tokens.
In the previous section, we have demonstrated the effectiveness of the selection of the masking tokens, such as the whole word masking or N-gram masking, etc.
Now we are going to investigate how the replacement of the selected tokens affects the performance of the pre-trained language models.
In order to investigate this problem, we plot the CMRC 2018 and DRCD performance at different pre-training steps.
Specifically, we follow the original masking percentage 15\% of the input sequence, in which 10\% masked tokens remain the same.
In terms of the remaining 90\% masked tokens, we classify them into four categories.
\begin{itemize}
	\item {\bf MacBERT}: 80\% tokens replaced into their similar words, and 10\% replaced into random words. 
	\item {\bf Random Replace}: 90\% tokens replaced into random words.
	\item {\bf Partial Mask}: original BERT implementation, with 80\% tokens replaced into {\tt [MASK]} tokens, and 10\% replaced into random words.	
	\item {\bf All Mask}: 90\% tokens replaced with {\tt [MASK]} tokens.
\end{itemize}
  
\begin{figure}[t]
\centering
\subfigure{
\begin{minipage}[htbp]{0.95\linewidth}
\centering
\includegraphics[width=0.8\columnwidth]{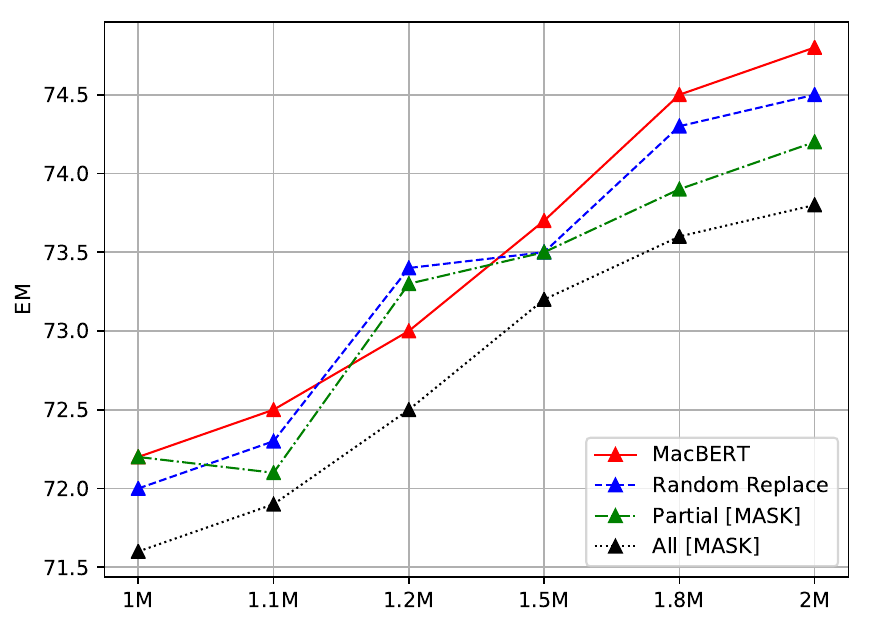} \\
\includegraphics[width=0.8\columnwidth]{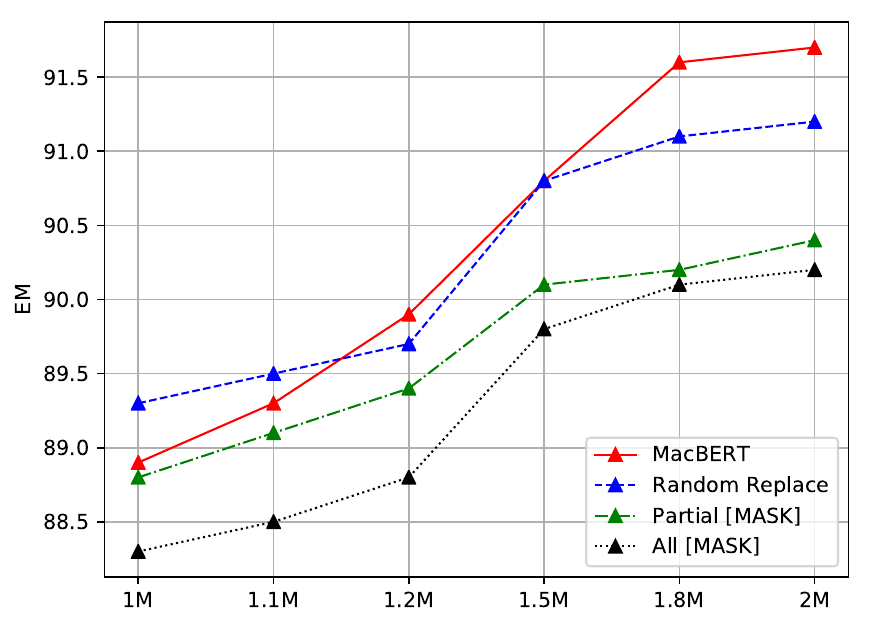}
\end{minipage}%
}%
\centering
  \caption{\label{discussion} Results of different MLM tasks on CMRC 2018 and DRCD.}
\end{figure}

We only plot the steps from 1M to 2M to show stabler results than the first 1M steps.
The results are depicted in Figure \ref{discussion}.

The pre-training models that rely on mostly using {\tt [MASK]} for masking purposes (i.e., partial mask and all mask) result in worse performances, indicating that the discrepancy of the pre-training and fine-tuning is an actual problem that affects the overall performance.
Among which, we also noticed that if we do not leave 10\% as original tokens (i.e., identity projection), there is also a consistent decline, indicating that masking with {\tt [MASK]} token is less robust and vulnerable to the absence of identity projection for negative sample training.

To our surprise, a quick fix, that is to abandon the {\tt [MASK]} token completely and replace all 90\% masked tokens into random words, yields consistent improvements over {\tt [MASK]}-dependent masking strategies. 
This also strengthens the claims that the original masking method that relies on the {\tt [MASK]} token, which never appears in the fine-tuning task, resulting in a discrepancy and worse performance. 
Also, using random words rather than the artificial token {\tt [MASK]} could improve the de-noising ability of the pre-trained model, which might also be a possible reason.
To make this more delicate, in this paper, we propose to use similar words for masking purposes, instead of randomly pick a word from the vocabulary, as random words are not fit in the context and may break the naturalness of the language model learning, as traditional N-gram language model is based on natural sentence rather than a manipulated influent sentence.
However, if we use similar words for masking purposes, the fluency of the sentence is much better than using random words, and the whole task transforms into a grammar correction task, which is much more natural and without the discrepancy of the pre-training and fine-tuning stage.
From the figure, we can see that the MacBERT yields the best performance among the four variants, which verifies our assumptions.

%%%%%%%%%%%%%%
\subsection{Analyses on Chinese Spell Check}
\begin{figure}[htp]
  \centering
  \subfigure[Detection-level]{\includegraphics[width=0.8\columnwidth]{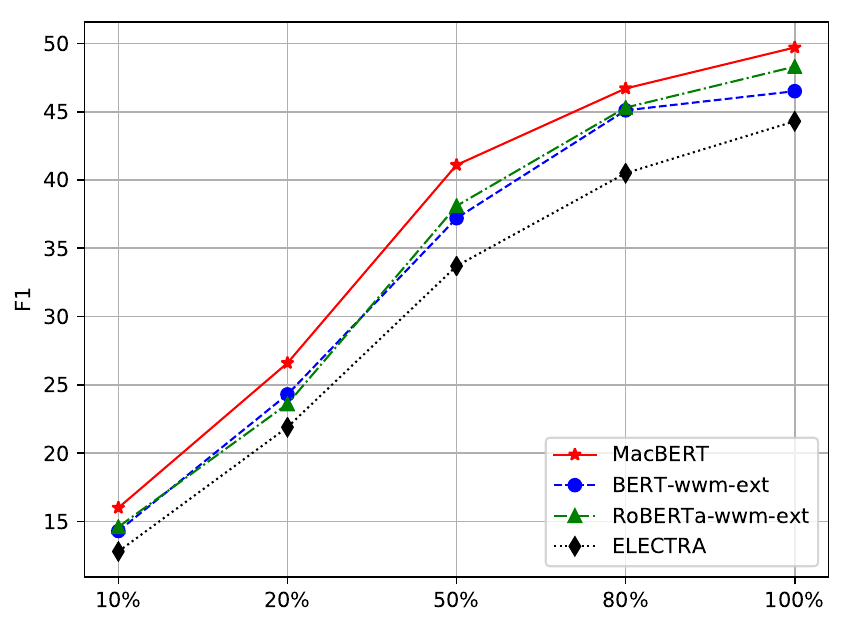}} \\
  \subfigure[Correction-level]{\includegraphics[width=0.8\columnwidth]{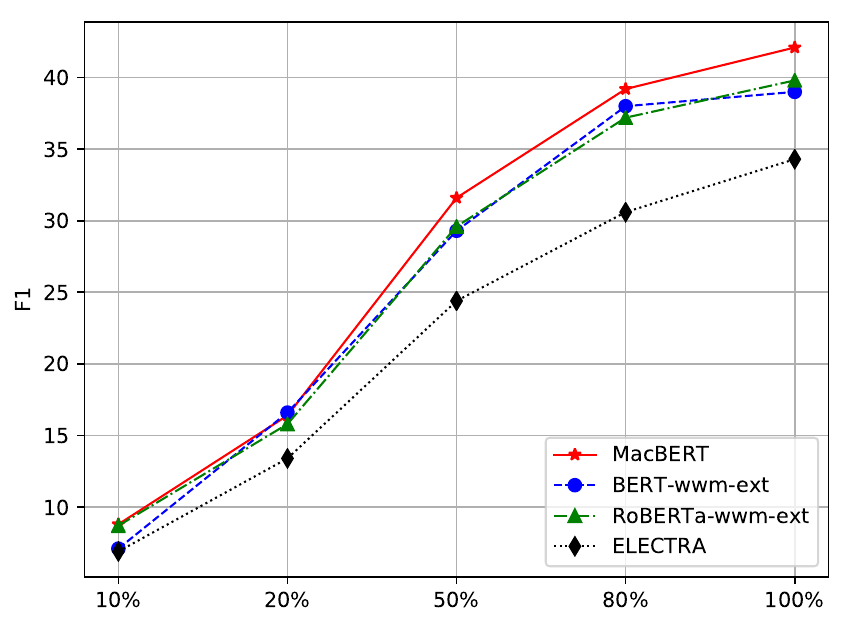}}
   \caption{\label{sighan-analysis} Results of using different percentage of SIGHAN-15 training data.}
\end{figure}

MacBERT introduces `MLM as correction' tasks, which is similar to the actual grammar or spell error correction tasks.
We perform additional experiments on Chinese Spell Check tasks. 
We use SIGHAN-15 \cite{tseng-etal-2015-introduction} dataset to explore the effect of different pre-trained language models when using different percentages of training data. 
SIGHAN-15 consists of a training set of 3.1K instances and a test set of 1.1K instances.
We compare BERT-wwm-ext, RoBERTa-wwm-ext, ELECTRA-base, and MacBERT-base in this experiment, as they share the same pre-training data.
We fine-tune each model five times and plot the figures with averaged F1 (sentence-level).
We use a universal learning rate of 5e-5 and train 5 epochs with a batch size of 64.
The results are shown in Figure \ref{sighan-analysis}, including detection-level and correction-level scores.

As we can see that our MacBERT yields consistent improvements over others when using different percentages of the training data, indicating that our approach is effective and scalable.
We notice that ELECTRA does not perform well on this task.
Especially, the gap between ELECTRA and others on the correction-level results are relatively larger than that in the detection-level.
ELECTRA uses replaced token detection (RTD) task for training the discriminator (which will be used for fine-tuning).
However, the RTD task only needs to identify whether the input tokens are altered without predicting the original token, which we think is quite simple.
On the contrary, MLM and Mac objectives require identify-and-correction at the same time.
By comparing MLM and Mac, our MacBERT alleviates the discrepancy of pre-training and fine-tuning issues, which yields another significant gain.

We note that though the Mac task is similar to the spell check task, we only use synonyms for replacement, which is only a small proportion in real spell check tasks. 
This could explain why our model does not yield larger improvement over others when there is fewer training data available.

%%%%%%%%%%%%%%%%%%%%%%%%%%%%%%%%%%%%%%%%
\section{Conclusion}
In this paper, we revisit pre-trained language models in Chinese to see if the techniques in these state-of-the-art models generalize well in a different language other than English only.
We created Chinese pre-trained language model series and proposed a new model called MacBERT, which modifies the masked language model (MLM) task as a language correction manner and mitigates the discrepancy of the pre-training and fine-tuning stage.
Extensive experiments are conducted on various Chinese NLP datasets, and the results show that the proposed MacBERT could give significant gains in most of the tasks, and detailed ablations show that more focus should be made on the MLM task rather than the NSP task and its variants, as we found that NSP-like task does not show a landslide advantage over one another.
With the release of the Chinese pre-trained language model series, we hope it will further accelerate the natural language processing in our research community.

In the future, we would like to investigate an effective way to determine the masking ratios instead of heuristic ones to further improve the performance of the pre-trained language models.
Also, we would like to design more effective language modeling approaches to further exploit large-scale unsupervised data.

\section*{Acknowledgment}
We would like to thank all anonymous reviewers and editors for their thorough reviewing and providing constructive comments to improve our paper. 
The first author was partially supported by the Google TPU Research Cloud (TRC) program for Cloud TPU access.

\ifCLASSOPTIONcaptionsoff
  \newpage
\fi



\bibliographystyle{IEEEtran}
\bibliography{IEEEexample.bib}


\begin{IEEEbiography}[{\includegraphics[width=1in,height=1.25in,clip,keepaspectratio]{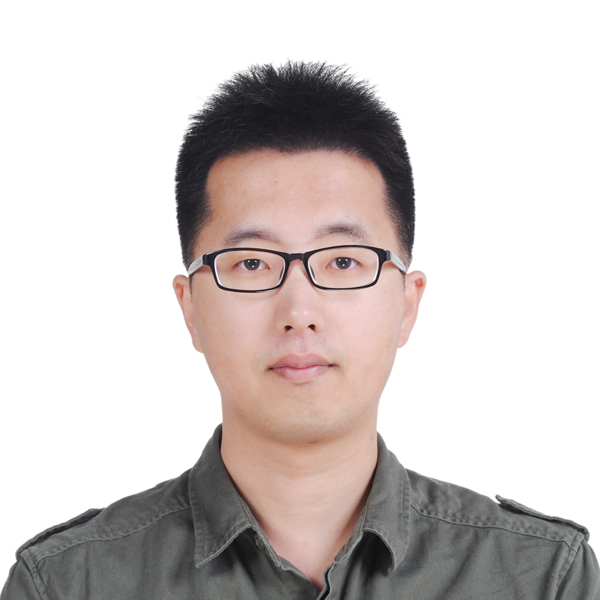}}]{Yiming Cui}
is a principal researcher at Joint Laboratory of HIT and iFLYTEK Research (HFL).
He received M.S. and B.S. degrees and is currently pursuing a doctoral degree at Harbin Institute of Technology.
His main research interests include Machine Reading Comprehension, Question Answering, and Pre-trained Language Model, etc. 
He has published more than 20 papers in top conferences, such as in ACL, EMNLP, AAAI, COLING, NAACL, etc.
He serves as a senior member of China Computer Federation (CCF).
\end{IEEEbiography}

\begin{IEEEbiography}[{\includegraphics[width=1in,height=1.25in,clip,keepaspectratio]{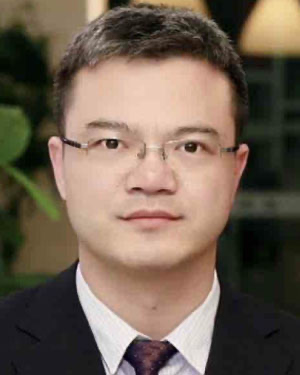}}]{Wanxiang Che}
is a professor of School of Computer Science and Technology, Harbin Institute of Technology. He is the vice director of Research Center for Social Computing and Information Retrieval. He is a young scholar of “Heilongjiang Scholar” and a visiting scholar of Stanford University. He is currently the vice director and secretary-general of the Computational Linguistics Professional Committee of CIPS and a CCF senior member. He achieved the AAAI 2013 Outstanding Paper Honorable Mention Award. His Language Technology Platform (LTP) has been shared by over 600 organizations, and authorized to Baidu, Tencent, and so on. 
\end{IEEEbiography}

\begin{IEEEbiography}[{\includegraphics[width=1in,height=1.25in,clip,keepaspectratio]{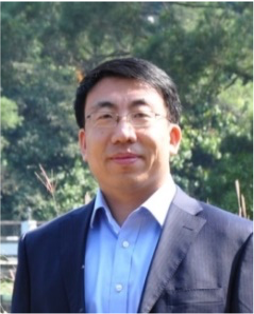}}]{Ting Liu}
received his Ph.D. degree in 1998 from the Department of Computer Science, Harbin Institute of Technology. 
He is a Full Professor and Director of the Department of Computer Science, from Harbin Institute of Technology. 
His research interests include information retrieval, natural language processing, and social media analysis.
He has published hundreds of papers with over 20,000 citations.
\end{IEEEbiography}

\begin{IEEEbiography}[{\includegraphics[width=1in,height=1.25in,clip,keepaspectratio]{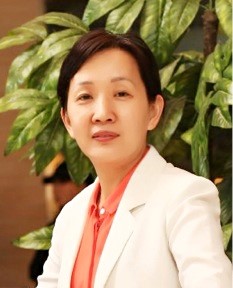}}]{Bing Qin}
is a full professor and doctoral supervisor of School of computer science, Harbin Institute of Technology. 
She is also the Director of Research Center for Social Computing and Information Retrieval (HIT-SCIR) from Harbin Institute of Technology.
Her main research directions include natural language processing, information extraction, text mining, emotion analysis, etc. She has published more than 80 papers in top conferences, such as ACL, COLING, EMNLP, IEEE TKDE, IEEE TASLP, etc. 
\end{IEEEbiography}

\begin{IEEEbiography}[{\includegraphics[width=1in,height=1.25in,clip,keepaspectratio]{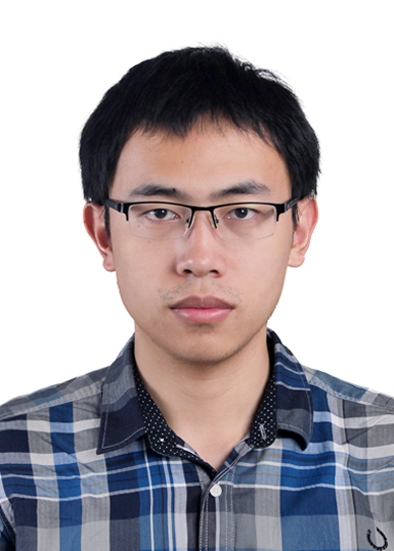}}]{Ziqing Yang}
is a researcher at Joint Laboratory of HIT and iFLYTEK Research (HFL). He received his B.S degree from Wuhan University in 2010 and received his Ph.D. degree in Physics from University of Chinese Academy of Sciences in 2017. He has a broad interest in machine learning and natural language processing, including machine reading comprehension, knowledge distillation for NLP and general machine learning method for NLP. He has published several top-tier conference papers, including ACL, COLING, etc.
\end{IEEEbiography}

%\newpage


%\vfill


\end{CJK*}
\end{document}